\documentclass{article}

\usepackage{amssymb}
\usepackage{amsmath}
\usepackage{amsthm}
\usepackage{booktabs}
\usepackage{catchfile}
\usepackage{colonequals}
\usepackage{color}
\usepackage{epigraph}
\usepackage{float}
\usepackage{graphicx}
\usepackage{hyperref}
\usepackage{listings}
\usepackage{pgfplots}
\usepackage{siunitx}
\usepackage{subcaption}
\usepackage{standalone}
\usepackage{tikz}
\usepackage{tikzscale}
\usepackage{times}
\usepackage[disable]{todonotes} % Pass 'disable' as option to hide todos.
\usetikzlibrary{calc, math}
\hypersetup{colorlinks = true, urlcolor = blue, linkcolor = black}
\usepgfplotslibrary{external}
\newtheorem{theorem}{Theorem}
\newtheorem{corollary}[theorem]{Corollary}

\theoremstyle{remark}

\DeclareMathOperator{\diag}{diag}

\allowdisplaybreaks

\lstdefinestyle{pystyle}{
  basicstyle=\footnotesize
  }

\title{Intelligent Matrix Exponentiation}

\author{Thomas Fischbacher \and
        Iulia M. Comsa \and
        Krzysztof Potempa \and
        Moritz Firsching \and
        Luca Versari \and
        Jyrki Alakuijala}
\date{\vspace{-2em}}
\begin{document}
\maketitle
\begin{center}
{\em Google Research\\
  Brandschenkestrasse 110, 8002 Z\"urich, Switzerland\\
}
{\small \texttt{\{tfish,iuliacomsa,dickstra,firsching,veluca,jyrki\}@google.com}}

\end{center}
\vskip 0.3in

\begin{abstract}
  \noindent We present a novel machine learning architecture
  that uses the exponential of a single input-dependent matrix as its only
  nonlinearity. The mathematical simplicity of this architecture allows a
  detailed analysis of its behaviour, providing robustness guarantees via
  Lipschitz bounds.  Despite its simplicity, a single matrix exponential layer
  already provides universal approximation properties and can learn fundamental
  functions of the input, such as periodic functions or multivariate
  polynomials.  This architecture outperforms other general-purpose
  architectures on benchmark problems, including CIFAR-10, using substantially
  fewer parameters.
\end{abstract}

\section{Introduction}
\label{sec:Introduction}

Deep neural networks (DNNs) synthesize highly complex functions by composing a
large number of neuronal units, each featuring a basic and usually
$1$-dimensional nonlinear activation function~$f : \mathbb{R}^1 \to
\mathbb{R}^1$.  While highly successful in practice, this approach also has
disadvantages.  In a conventional DNN, any two activations only ever get
combined through summation. This means that such a network requires an
increasing number of parameters to express more complex functions even as
simple as multiplication. This approach of composing simple functions does not
generalize well outside the boundaries of the training data.

An alternative to the composition of many 1-dimensional functions is using a
simple higher-dimensional nonlinear function~$f: \mathbb{R}^m\to\mathbb{R}^n$.
A single multidimensional nonlinearity may be desirable because it could express
more complex relationships between input features with potentially fewer
parameters and fewer mathematical operations.

The matrix exponential stands out as a promising but overlooked candidate for a
higher-dimensional nonlinearity that may be used as a building block for machine
learning models. The matrix exponential is a smooth function governed by a
relatively simple equation that yields desirable mathematical properties.
It has applications in solving linear differential equations
and plays a prominent role in the theory of Lie groups, an algebraic
structure widely used throughout many branches of mathematics and science.

We propose a novel ML architecture for supervised learning whose core element
is \emph{a single layer} (henceforth referred to as ``M-layer''), that computes
\emph{a single matrix exponential}, where the matrix to be exponentiated is an
affine function of the input features. We show that the M-layer has universal
approximator properties and allows closed-form per-example bounds for
robustness. We demonstrate the ability of this architecture to learn
multivariate polynomials, such as matrix determinants, and to generalize
periodic functions beyond the domain of the input without any feature
engineering. Furthermore, the M-layer achieves results comparable to
recently-proposed non-specialized architectures on image recognition datasets.
We provide open-source TensorFlow code that implements the M-layer:
\url{https://github.com/google-research/google-research/tree/master/m_layer}.

\section{Related Work}
\label{sec:RelatedWork}

Neuronal units with more complex activation functions have been proposed. One
such example are sigma-pi units \cite{rumelhart1986pdp2}, whose activation
function is the weighted sum of products of its inputs.  More recently, neural
arithmetic logic units have been introduced \cite{trask2018}, which can combine
inputs using multiple arithmetic operators and generalize outside the domain of
the training data.  In contrast with these architectures, the M-layer is not
based on neuronal units with multiple inputs, but uses a single matrix
exponential as its nonlinear mapping function. Through the matrix exponential,
the M-layer can easily learn mathematical operations more complex than
addition, but with simpler architecture. In fact, as shown in
Section~\ref{sec:FeatureCrossesUniversalApproximation}, the M-layer can be
regarded as a generalized sigma-pi network with built-in architecture search,
in the sense that it learns by itself which arithmetic graph should be used for
the computation.

Architectures with higher-dimensional nonlinearities are also already used.
The softmax function is an example for a widely-used such nonlinear activation
function that solves a specific problem, typically in the final layer of
classifiers. Like the M-layer, it has extra mathematical structure.  For
example, a permutation of the softmax inputs produces a corresponding
permutation of the outputs.  Maxout networks also act on multiple units and
have been successful in combination with dropout \cite{Goodfellow2013}. In
radial basis networks \cite{park1991}, each hidden unit computes a nonlinear
function of the distance between its own learned centroid and a single point
represented by a vector of input coordinates.  Capsule networks
\cite{sabour2017} are another recent example of multidimensional
nonlinearities. Similarly, the M-layer uses the matrix exponential as a single
high-dimensional nonlinearity, therefore creating additional mathematical
structure that potentially allows solving problems using fewer parameters than
compositional architectures.

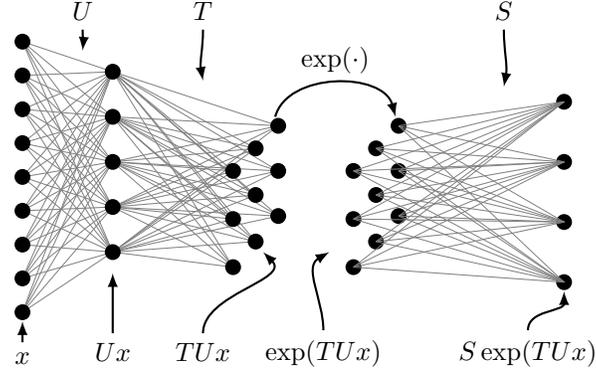
\begin{figure}
\centering
\begin{tikzpicture}[ultra thin,
                    gray,
                    scale=0.4,
                    every circle node/.style=
                        {fill=black,
                         inner sep=0pt,
                         minimum size=6pt}]
  %Choose x-coordinate, top, bottom and number of points - 1 for layers.
  \tikzmath{\x0 = -2;
            \b0 = 0;
            \t0 = 9;
            \n0 = 8;
            \x1 = 1;
            \b1 = 2;
            \t1 = 8;
            \n1 = 4;
            \n2 = 2;
            \x4 = 16;
            \b4 = 1;
            \t4 = 7;
            \n4 = 3;
            }
  \coordinate (A0) at (\x0, \b0);
  \coordinate (B0) at (\x0, \t0);
  \coordinate (A1) at (\x1, \b1);
  \coordinate (B1) at (\x1, \t1);
  % Draw the edges and nodes.
  \foreach \y in {0,..., \n0} {
    \foreach \yy in {0,...,\n1} {
      \draw ($(A0)!\y/\n0!(B0)$) -- ($(A1)!\yy/\n1!(B1)$);
    }
  }
  \coordinate (A) at (5,1.5);
  \coordinate (B) at (5,4.7);
  \coordinate (C) at (6.5,3.2);
  % Without perpsective (D) would be at ($(B) + (C) - (A)$).
  \coordinate (D) at (6.5, 6.2);

  \foreach \x in {0,...,\n2} {
    \foreach \y in {0,...,\n2} {
      \foreach \yy in {0,...,\n1} {
        \draw ($(A1)!\yy/\n1!(B1)$) --  ($ ($(A)!\x/\n2!(B)$)!\y/\n2!($(C)!\x/\n2!(D)$) $);
      }
    }
  }

  \coordinate (E) at ($(A) + (4,0)$);
  \coordinate (F) at ($(B) + (4,0)$);
  \coordinate (G) at ($(C) + (4,0)$);
  \coordinate (H) at ($(D) + (4,0)$);
  \coordinate (A4) at (\x4, \b4);
  \coordinate (B4) at (\x4, \t4);

  \foreach \yy in {0,...,\n4} {
    \node at ($(A4)!\yy/\n4!(B4)$) [circle] {};
  }
  \foreach \x in {0,...,\n2} {
    \foreach \y in {0,...,\n2} {
     \node at  ($ ($(E)!\x/\n2!(F)$)!\y/\n2!($(G)!\x/\n2!(H)$) $) [circle] {};
     \node at ($ ($(A)!\x/\n2!(B)$)!\y/\n2!($(C)!\x/\n2!(D)$) $) [circle] {};
     }
   }
  \foreach \x in {0,...,\n2} {
    \foreach \y in {0,...,\n2} {
      \foreach \yy in {0,...,\n4} {
        \draw ($ ($(E)!\x/\n2!(F)$)!\y/\n2!($(G)!\x/\n2!(H)$) $) -- ($(A4)!\yy/\n4!(B4)$);
      }
    }
  }
  \foreach \y in {0,...,\n0} {
    \node at ($(A0)!\y/\n0!(B0)$) [circle] {};
  }
  \foreach \y in {0,...,\n1} {
    \node at ($(A1)!\y/\n1!(B1)$) [circle] {};
  }
  % Place the labels.
  \tikzset{every path/.style={->,black,thick}}

  \draw[-latex](0,9.4) node[above]{$U$} to[out=270,in=85] (0,8.7);
  \draw[-latex](4,9.4) node[above]{$T$} to[out=270,in=100] (4,7.7);
  \draw[-latex] (6.4, 6.5) node[above right]{} to[out=75, in=105]
                node[above, swap]{$\exp(\cdot)$}(10.4,6.5);
  \draw[-latex](14,9.4) node[above]{$S$} to[out=270,in=80] (14,7.5);

  \draw[-latex,thick](\x0,-1) node[below]{$x$} to[out=90,in=270] (\x0, -.3);
  \draw[-latex,thick](\x1,-.7) node[below]{$Ux$} to[out=90,in=270] (\x1,1.3);
  \draw[-latex,thick](4,-.7) node[below]{$TUx$} to[out=90,in=315] (6,2);
  \draw[-latex,thick](8,-.6) node[below]{$\exp(TUx)$} to[out=90,in=210] (8.2,2);
  \draw[-latex,thick](\x4-1.2,-.6) node[below]{$S\exp(TUx)$}
                              to[out=90,in=270] (\x4,.7);
\end{tikzpicture}
  \caption{Schematic diagram of the M-layer architecture.}
  \label{fig:mel}
\end{figure}

Matrix exponentiation has a natural alternative interpretation in terms of an
ordinary differential equation (ODE). As such, the M-layer can be compared to
other novel ODE-related architectures that have been proposed recently.  In
particular, neural ordinary differential equations (NODE) \cite{chen2018} and
their augmented extensions (ANODE) \cite{dupont2019} have recently received
attention. We discuss this in Section~\ref{sec:Interpretation}.

Existing approaches to certifying the robustness of neural networks can be
split into two different categories. Some approaches \cite{peck2017lower}
mathematically analyze a network layer by layer, providing bounds on the
robustness of each layer, that then get multiplied together. This kind of
approach tends to give fairly loose bounds, due to the inherent tightness loss
from composing upper bounds. Other approaches \cite{singh2018boosting,
singh2019abstract} use abstract interpretation on the evaluation of the network
to provide empirical robustness bounds. In contrast, using the fact that the
M-layer architecture has a single layer, in Section~\ref{sec:Robustness} we
obtain a direct bound on the robustness on the whole network by analyzing the
explicit formulation of the computation.

\section{Architecture}
\label{sec:Architecture} We start this section by refreshing the definition of
the matrix exponential. We then define the proposed M-layer model and explain
its ability to learn particular functions such as polynomials and periodic
functions. Finally, we provide closed-form per-example robustness guarantees.

\subsection{Matrix Exponentiation}
\label{sec:MatrixExponentiation}

The exponential of a square matrix $M$ is defined as:
\begin{equation}
\label{eq:exp}
  \exp(M) = \sum_{k=0}^{\infty} \frac{1}{k!} M^k
\end{equation}

The matrix power~$M^k$ is defined inductively as~${M^0=I}$, ${M^{k+1}=M\cdot
M^k}$, using the associativity of the matrix product; it is not an
element-wise matrix operation.

Note that the expansion of $\exp(M)$ in Eq.~(\ref{eq:exp}) is finite for
nilpotent matrices. A matrix $M$ is called \emph{nilpotent} if there exists a
positive integer $k$ such that $M^k = 0$. Strictly upper triangular matrices
are a canonical example.

Multiple algorithms for computing the matrix exponential efficiently have been
proposed \cite{moler2003}. TensorFlow implements \texttt{tf.linalg.expm}
using the scaling and squaring method combined with the Pad\'e approximation
\cite{higham2005}.

\subsection{M-Layer Definition}
\label{sec:MLayerDefinition}

At the core of the proposed architecture is an M-layer that computes
a \emph{single} matrix exponential, where the matrix to be exponentiated is an
affine function of all of the input features. In other words, an M-layer
replaces \emph{an entire stack of hidden layers} in a DNN.

Figure~\ref{fig:mel} shows a diagram of the proposed architecture.  We
exemplify the architecture as applied to a standard image recognition dataset,
but we note that this formulation is applicable to any other type of problem by
adapting the relevant input indices. In the following equations, generalized
Einstein summation is performed over all right-hand side indices not seen on
the left-hand side. This operation is implemented in TensorFlow by
\texttt{tf.einsum}.

Consider an example input image, encoded as a $3$-index
array $X_{yxc}$, where $y$, $x$ and $c$ are the row index, column index and
color channel index, respectively. The matrix $M$ to be exponentiated is
obtained as follows, using the trainable parameters $\tilde T_{ajk}$,
$\tilde U_{axyc}$ and $\tilde B_{jk}$:
\begin{equation}
\label{eq:WeightedMatrix}
  M = \tilde B_{jk} + \tilde T_{ajk} \tilde U_{ayxc} X_{yxc}
\end{equation}
$X$ is first projected linearly to a $d$-dimensional latent
feature embedding space by $\tilde U_{ayxc}$. Then, the $3$-index tensor
$\tilde T_{ajk}$ maps each such latent feature to an $n\times n$ matrix.
Finally, a bias matrix $\tilde B_{jk}$ is added to the feature-weighted
sum of matrices. The result is a matrix indexed by row and column indices
$j$ and $k$.

We remark that it is possible to contract the tensors $\tilde T$ and $\tilde U$
in order to simplify the architecture formula, but partial tensor factorization
provides regularization by reducing the parameter count.

An output $p_m$ is obtained as follows, using the trainable parameters
$\tilde S_{mjk}$ and $\tilde V_{m}$:
\begin{equation}
\label{eq:MLayerOutput}
  p_m = \tilde V_{m} + \tilde S_{mjk} \exp(M)_{jk}
\end{equation}
The matrix $\exp(M)$, indexed by row and column indices $j$ and $k$ in the
same way as $M$, is projected linearly by the $3$-index tensor~$\tilde S_{mjk}$,
to obtain a $h$-dimensional output vector. The bias-vector~$\tilde V_{m}$
turns this linear mapping into an affine mapping. The resulting vector may
be interpreted as accumulated per-class evidence and, if desired, may then
be mapped to a vector of probabilities via softmax.

Training is done conventionally, by minimizing a loss function such as the
$L_2$ norm or the cross-entropy with softmax, using backpropagation
through matrix exponentiation.

The nonlinearity of the M-layer architecture is provided by the
$\mathbb{R}^d\to\mathbb{R}^h$ mapping
$v\mapsto \tilde V_{m}+\tilde S_{mjk}\exp(M)_{jk}$.
The count of trainable parameters of this component is $dn^2 + n^2 + n^2h + h$.
This count comes from summing the dimensions of $\tilde T_{ajk}$, $\tilde
B_{jk}$, $\tilde S_{mjk}$, and $\tilde V_{m}$, respectively. We note that this
architecture has some redundancy in its parameters, as one can freely multiply
the $T$ and $U$ tensors by a $d \times d$ real matrix and, respectively, its
inverse, while preserving the computed function. Similarly, it is possible to
multiply each of the $n \times n$ parts of the tensors $\tilde T$ and $\tilde
S$, as well as $B$, by both an $n\times n$ matrix and its inverse. In other
words, any pair of real invertible matrices of sizes $d\times d$ and $n\times
n$ can be used to produce a new parametrization that still computes the same
function.

\subsection{Feature Crosses and Universal Approximation}
\label{sec:FeatureCrossesUniversalApproximation}

A key property of the M-layer is its ability to generate arbitrary
exponential-polynomial combinations of the input features.  For
classification problems, M-layer architectures are a superset of
multivariate polynomial classifiers, where the matrix size constrains
the complexity of the polynomial while at the same time not uniformly
constraining its degree. In other words, simple multivariate polynomials of high
degree compete against complex multivariate polynomials of low degree.

We provide a universal approximator proof for the M-layer in the Supplementary
Material, which relies on its ability to express any multivariate polynomial in
the input features if a sufficiently large matrix size is used.  We provide
here an example that illustrates how feature crosses can be generated through
the matrix exponential.

Consider a dataset with the feature vector $(\phi_0, \phi_1, \phi_2)$
given by the $\tilde U \cdot x$ tensor contraction, where the relevant
quantities for the final classification of an example are assumed to be
$\phi_0$, $\phi_1$, $\phi_2$, $\phi_0 \phi_1$, and $\phi_1 \phi_2^2$.
To learn this dataset, we look for an exponentiated matrix that makes
precisely these quantities available to be weighted by the
trainable tensor $\tilde S$. To do this, we define three
$7 \times 7$ matrices $T_{0jk}$, $T_{1jk}$, and $T_{2jk}$ as
$T_{001} = T_{102} = T_{203} = 1$, $T_{024} = T_{225} = 2$, $T_{256} = 3$,
and $0$ otherwise.
We then define the matrix $M$ as:
\[
    M = \phi_0 T_0 + \phi_1 T_1 + \phi_2 T_2 =
     \left(\begin{smallmatrix}
0 & \phi_{0} & \phi_{1} & \phi_{2} & 0 & 0 & 0 \\
0 & 0 & 0 & 0 & 0 & 0 & 0 \\
0 & 0 & 0 & 0 & 2\phi_{0} & 2\phi_{2} & 0 \\
0 & 0 & 0 & 0 & 0 & 0 & 0 \\
0 & 0 & 0 & 0 & 0 & 0 & 0 \\
0 & 0 & 0 & 0 & 0 & 0 & 3\phi_{2} \\
0 & 0 & 0 & 0 & 0 & 0 & 0
\end{smallmatrix}\right)
\]
Note that $M$ is nilpotent, as $M^4=0$. Therefore, we obtain the following
matrix exponential, which contains the desired quantities in its leading row:
\begin{align*}
  \exp(M) & = I + M + \frac{1}{2}M^2 + \frac{1}{6}M^3 = \\
  & = \left(\begin{smallmatrix}
1 & \phi_{0} & \phi_{1} & \phi_{2} & \phi_{0} \phi_{1} & \phi_{1} \phi_{2} & \phi_{1} \phi_{2}^{2} \\
0 & 1 & 0 & 0 & 0 & 0 & 0 \\
0 & 0 & 1 & 0 & 2\phi_{0} & 2\phi_{2} & 3\phi_{2}^{2} \\
0 & 0 & 0 & 1 & 0 & 0 & 0 \\
0 & 0 & 0 & 0 & 1 & 0 & 0 \\
0 & 0 & 0 & 0 & 0 & 1 & 3\phi_{2} \\
0 & 0 & 0 & 0 & 0 & 0 & 1
\end{smallmatrix}\right)
\end{align*}
The same technique can be employed to encode any polynomial in the
input features using a $n\times n$ matrix, where~$n$ is one unit larger than
the total number of features plus the intermediate and final products
that need to be computed. The matrix size can be seen as regulating
the total capacity of the model for computing different feature crosses.

With this intuition, one can read the matrix as a ``circuit breadboard''
for wiring up arbitrary polynomials. When evaluated on features that only
take values $0$ and $1$, any Boolean logic function can be expressed.

\subsection{Feature Periodicity}
\label{sec:FeaturePeriodicity}

While the M-layer is able to express a wide range of functions using the
exponential of nilpotent matrices, non-nilpotent matrices can bring additional
utility. One possible application of non-nilpotent matrices is learning the
periodicity of input features. This is a problem where conventional DNNs
struggle, as they cannot naturally generalize beyond the distribution
of the training data. Here we illustrate how matrix exponentials can naturally
fit periodic dependency on input features, without requiring an explicit
specification of the periodic nature of the data.

Consider the matrix 
$M_r = \left(\begin{smallmatrix}0&-\omega\\\omega&0\end{smallmatrix}\right)$.
We have $\exp(t M_r) = \left(\begin{smallmatrix}\cos \omega t&-\sin \omega t
\\ \sin \omega t& \phantom+\cos \omega t\end{smallmatrix}\right)$,
which is a 2d rotation by an angle of $\omega t$ and thus periodic in $t$
with period $2\pi/\omega$. This setup can fit functions that have an
arbitrary period. Moreover, this representation of periodicity naturally
extrapolates well when going beyond the range of the initial numerical data.

%We remark that allowing matrices over complex numbers or quaternions would not
%give the architecture any additional expressive power. Indeed, every $n\times
%n$ complex (respectively quaternionic) can be equivalently represented by a
%real $2n \times 2n$ (respectively $4n\times4n$) matrix while preserving the
%exponential map.

\subsection{Connection to Lie Groups}
The M-layer has a natural connection to Lie groups. Lie groups can be thought
of as a model of continuous symmetries of a system such as rotations. There is
a large body of mathematical theory and tools available to study the structure
and properties of Lie groups \cite{gilmore2008, gilmore2012}, which may
ultimately also help for model interpretability.

Every Lie group has associated a Lie algebra, which can be understood as the
space of the small perturbations with which it is possible to generate the
elements of the Lie group. As an example, the set of rotations of
$3$-dimensional space forms a Lie group; the corresponding algebra can be
understood as the set of rotation axes in $3$ dimensions. Lie groups and
algebras can be represented using matrices, and by computing a matrix
exponential one can map elements of the algebra to elements of the group.

In the M-layer architecture, the role of the $3$-index tensor~$\tilde T$ is to
form a matrix whose entries are affine functions of the input features. The
matrices that compose $\tilde T$ can be thought of as generators of a Lie
algebra. Building $M$ corresponds to selecting a Lie algebra element.  Matrix
exponentiation then computes the corresponding Lie group element.

% Maybe in follow-up work.
%We remark that the matrices that compose $\tilde T$ do not necessarily
%form a basis of the Lie algebra, as the definition of basis requires it to be
%closed under the operation of taking commutators, which does not in general
%hold for the matrices of $\tilde T$.

As rotations are periodic and one of the simplest forms of continuous
symmetries, this perspective is useful for understanding the ability of the
M-layer to learn periodicity in input features.

\subsection{Dynamical Systems Interpretation}
\label{sec:Interpretation} Recent work has proposed a dynamical systems
interpretation of some DNN architectures. The NODE architecture \cite{chen2018}
uses a nonlinear and not time-invariant ODE that is provided by trainable
neural units, and computes the time evolution of a vector that is constructed
from the input features.  This section discusses a similar interpretation of
the M-layer.

Consider an M-layer with $\tilde T$ defined as $\tilde T_{012}=\tilde
T_{120}=\tilde T_{201}=+1$, $\tilde T_{210}=\tilde T_{102}=\tilde T_{021}=-1$,
and $0$ otherwise, with $\tilde U$ as the $3 \times 3$ identity matrix, and
with $\tilde B = 0$. Given an input vector $a$, the corresponding matrix $M$
is then $\left(\begin{smallmatrix}
  0 & a_2 & -a_1 \\
  -a_2 & 0 & a_0 \\
  a_1 & -a_0 & 0 \\
\end{smallmatrix}\right)$. Plugging $M$ into the linear and time
invariant (LTI) ODE $d/dt\,Y(t)=MY(t)$, we can observe that the ODE describes a
rotation around the axis defined by $a$.  Moreover, a solution to this ODE is
given by~$Y(t)=\exp(tM)Y(0)$. Thus, by choosing $S_{mjk} = Y(0)_k$ if $m=j$ and
$0$ otherwise, the above M-layer can be understood as applying a
rotation with input dependent angular velocity to some basis vector
over a unit time interval.

More generally, we can consider the input features to provide affine
parameters that define a time-invariant linear ODE, and the output of
the M-layer to be an affine function of a vector that has evolved
under the ODE over a unit time interval. In contrast, the NODE
architecture uses a non-linear ODE that is not input dependent, which
gets applied to an input-dependent feature vector.

\label{sec:Results}
% here for twocolumn
\begin{figure*}
  \centering
  \begin{subfigure}[t]{0.5\textwidth}
    \input{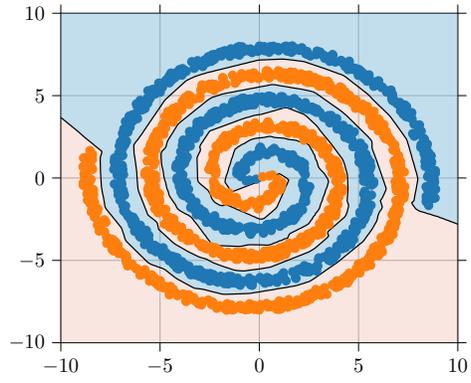}
    \caption{ReLU}
  \end{subfigure}
  \begin{subfigure}[t]{0.5\textwidth}
    \input{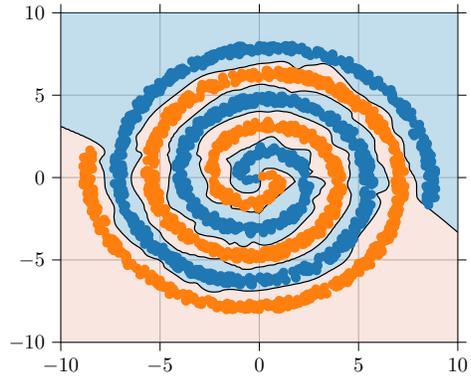}
    \caption{tanh}
  \end{subfigure}
  \begin{subfigure}[t]{0.5\textwidth}
    \input{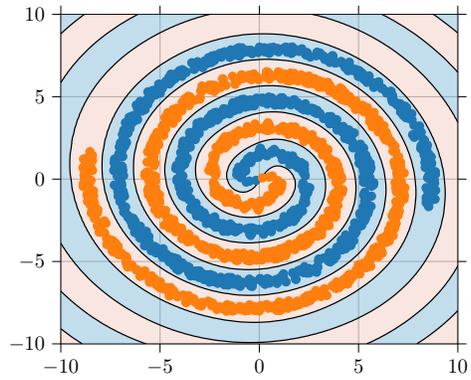}
    \caption{M-layer}
  \end{subfigure}
  \caption{Comparison of ReLU/tanh DNN and M-layer classification boundaries
  on a double spiral (``Swiss roll'') classification task.\label{fig:spirals}}
\end{figure*}

\subsection{Certified Robustness}
\label{sec:Robustness}

We show that the mathematical structure of the M-layer allows a novel
proof technique to produce closed-form expressions for guaranteed robustness
bounds.

For any matrix norm $\lVert\cdot\rVert$, we have \cite{horn1994}:
\[\lVert\exp(X + Y) - \exp(X)\rVert\leq \lVert Y\rVert\exp(\lVert Y \rVert)\exp(\lVert X \rVert)\]
We also make use of the fact that $\lVert M \rVert_F \le \sqrt{n} \lVert M
\rVert_2$ for any $n \times n$ matrix, where $\lVert \cdot \rVert_F$ is the
Frobenius norm and $\lVert \cdot \rVert_2$ is the $2$-norm of a matrix. We
recall that the Frobenius norm of a matrix is equivalent to the $2$-norm of the
vector formed from the matrix entries.

Let $M$ be the matrix to be exponentiated corresponding to a given input
example $x$, and let $M'$ be the deviation to this matrix that corresponds to
an input deviation of $\tilde x$, i.e. $M + M'$ is the matrix corresponding to
input example $x + \tilde x$. Given that the mapping between
$x$ and $M$ is linear, there is a per-model constant $\delta_{in}$ such that
$\lVert M'\rVert_2 \le \delta_{in} \lVert \tilde x \rVert_{\infty}$.

The $2$-norm of the difference between the outputs can be bound as follows:
\begin{align*}
  \lVert \Delta_{o} \rVert_2 & \le&\hskip-0.1in& \lVert S\rVert_2 \lVert \exp(M+M') - \exp(M)\rVert_F \le \\
                             & \le&\hskip-0.1in& \sqrt{n} \lVert S\rVert_2 \lVert \exp(M+M') - \exp(M)\rVert_2 \le \\
                             & \le&\hskip-0.1in& \sqrt{n} \lVert S\rVert_2 \lVert M'\rVert_2\exp(\lVert M' \rVert_2)\exp(\lVert M \rVert_2) \le \\
                             & \le&\hskip-0.1in& \sqrt{n} \lVert S\rVert_2 \delta_{in} \lVert \tilde x\rVert_\infty
                                                 \exp(\delta_{in}\lVert \tilde x \rVert_\infty)\exp(\lVert M \rVert_2)
\end{align*}
where $\lVert S \rVert_2$ is computed by considering $S$ a $h \times n\cdot n$
rectangular matrix, and the first inequality follows from the fact that the
tensor multiplication by $S$ can be considered a matrix-vector multiplication
between $S$ and the result of matrix exponential seen as a $n \cdot n$
vector.

This inequality allows to compute the minimal $L_\infty$ change required in the
input given the difference between the amount of accumulated evidence between
the most likely class and other classes. Moreover, considering that $\lVert x
\rVert_\infty$ is bounded from above, for example by $1$ in the case of
CIFAR-10, we can obtain a Lipschitz bound by replacing the
$\exp(\delta_{in}\lVert \tilde x \rVert_\infty)$ term with a
$\exp(\delta_{in})$ term.

\section{Results}
In this section, we demonstrate the performance of the M-layer on multiple
benchmark tasks, in comparison with more traditional architectures.
We first investigate the shape of the classification boundaries in a
classic double spiral problem. We then show that the M-layer is able to
learn determinants of matrices up to size $5 \times 5$, periodic functions
in the presence of low noise, and image recognition datasets at a level
competitive with other non-specialized architectures. For CIFAR-10, we compare
the training times and robustness to those of traditional DNNs.

The following applies for all experiments below, unless otherwise stated.
DNN models are initialized using uniform Glorot
initialization \cite{glorot2010understanding}, while M-layer models are
initialized with normally distributed values with mean $0$ and $\sigma=0.05$.
To enhance training stability and model performance,
an activity regularization is performed on the output of the M-layer.
This is achieved by adding $\lambda \lVert \exp(M) \rVert^2_F$ to the loss
function with a value of $\lambda$ equal to $10^{-4}$. This value is chosen
because it performs best on the CIFAR-10 dataset from a choice of
$10^{-3}$, $5 \cdot 10^{-3}$, $10^{-4}$, $5\cdot 10^{-4}$, and $5\cdot 10^{-5}$.

\begin{figure*}
  \centering
  \begin{subfigure}[t]{0.6\textwidth}
    \input{periodic3-4.tikz}
    \caption{\scalebox{1}{$1.42 \cos(6\pi x+0.84) + 6.29 \cos(8\pi x+0.76)$}}
  \end{subfigure}
  \begin{subfigure}[t]{0.6\textwidth}
    \input{periodic4-7.tikz}
    \caption{\scalebox{1}{$1.55 \cos(8\pi x+0.26) + 5.07 \cos(14\pi x+0.81)$}}
  \end{subfigure}
  \begin{subfigure}[t]{0.6\textwidth}
    \input{periodic6-8.tikz}
    \caption{\scalebox{1}{$1.60 \cos(12\pi x+0.59) + 6.92 \cos(16\pi x+0.44)$}}
  \end{subfigure}
  \caption{Learning periodic functions with a DNN and an M-layer.  Each plot
  shows outputs from three separate models of each type.} \label{fig:periodic}
\end{figure*}

\subsection{Learning Double Spirals}

To compare the classification boundaries generated by the M-layer with those of
more traditional architectures, we train DNNs with ReLU and tanh activation
functions, as well as M-layers, using a double spiral (``Swiss roll'')
classification task as a toy problem.

The data consist of $2000$ randomly generated points along two spirals,
with coordinates in the $[-10, 10]$ range. Uniform random noise in the $[-0.5,
0.5]$ range is added to each input coordinate. As we are only
interested in the classification boundaries, no test or validation set is used.

The M-layer has a representation size $d=10$ and a matrix size $n=1$.
Each DNN has two hidden layers of size $20$.

A RMSprop optimizer is used to minimize the cross-entropy with softmax. The
M-layer is trained for $100$ epochs using a learning rate of $0.001$. The ReLU
DNN is trained for $1000$ epochs using a learning rate of $0.001$. The tanh
DNN is trained for $1000$ epochs using a learning rate of $0.01$. These values
are chosen in such a way that all networks achieved a perfect fit.

The resulting boundaries for the three models are shown in
Figure~\ref{fig:spirals}. They illustrate the distinctive ability of the
M-layer to extrapolate functions beyond the training domain.

\subsection{Learning Periodic Functions}
\label{sec:LearningPeriodicFunctions}

To assess the capacity of the M-layer architecture to learn and extrapolate
feature periodicity, we compare the performance of an M-layer and a DNN on
periodic functions obtained as the sum of two cosines.

The data is generated as follows. The frequencies of the
cosines are chosen as small integer multiples of $2\pi$ (from $3$ to $9$);
the amplitudes are randomly generated from the intervals $[1, 2]$ and $[5,10]$
respectively, and the phases are randomly generated in the $[0, \pi/3]$
range. Each model is trained on the $[0, 2]$ range and tested it on the $[2,
6]$ range, with a point spacing of $10^{-5}$.
Gaussian random noise with $\sigma = 10^{-4}$ is added to the target
value of each training sample. No activity regularization is used.

The M-layer uses a representation size $d=1$ and a matrix size $n=6$,
resulting in a trainable parameter count of $115$. Each cosine can be
represented by using a $2$-dimensional subspace; a matrix size of $4$
would thus be sufficient, but $6$ was chosen to show that an M-layer can learn
periodicity even when overparameterized.

In this experiment, the initialization of the bias and of $T_{0ij}$ is
performed by generating normally distributed numbers with $\sigma=0.01$ and
mean $-10$ for elements of the diagonal, and $0$ for all other elements.  The
coefficients of the mapping from input values to the embedding space are
initialized with normally distributed values with mean $0.1$ and $\sigma=0.05$.
This initialization is chosen in order to make it more likely for the initial
matrix $M$ to be exponentiated to have negative eigenvalues and therefore keep
outputs small.

The ReLU DNN is composed of two hidden layers with $50$ neurons each,
followed by one hidden layer with $10$ neurons, resulting in a trainable
parameter count of $3221$. The DNN was initialized using uniform Glorot
initialization \cite{glorot2010understanding}. As the objective of this
experiment is to demonstrate the ability to learn the periodicity of the
input without additional engineering, we do not consider DNNs with
special activation functions such as $\sin(x)$.

A RMSprop optimizer is used to minimize the following modified $L_2$ loss
function: if $f$ is the function computed by the network, $x$ the input of the
sample and $y$ the corresponding output, then the loss is given by $(f(x)-y)^2
+ \max(0, |f(2 x + 6)| - 100)^2$. In other words, very large values in
the $[6, 10]$ time range are punished.

M-layers are trained for $300$ epochs with learning rate $5\cdot10^{-3}$,
decay rate $10^{-5}$ and batch size $128$. DNNs are trained for $300$
epochs with learning rate $10^{-3}$, decay rate $10^{-6}$ and batch size $64$.
The hyperparameters are chosen by running multiple training steps with various
choices of learning rate ($10^{-2}$, $10^{-3}$, $10^{-4}$), decay rate
($10^{-3}$, $10^{-4}$, $10^{-5}$, $10^{-6}$), batch size ($64$ and $128$) and
number of epochs ($50$, $100$, $300$). For each model, the set of parameters
that provided the best $L_2$ loss on the training set is chosen.

Examples of functions learned by the M-layer and the DNN are shown in
Figure~\ref{fig:periodic}, which illustrates that, in contrast to the DNN, the
M-layer is able to extrapolate such functions.

\subsection{Learning Determinants}
\label{sec:LearningDeterminants}

To demonstrate the ability of the M-layer to learn polynomials, we
train an M-layer and a DNN to predict the determinant of
$3 \times 3$ and $5 \times 5$ matrices. We do not explicitly encode any
special property of the determinant, but rather employ it as an example
multivariate polynomial that can be learnt by the M-layer. We confirm
that we observe equivalent behavior for the matrix permanent.

Learning the determinant of a matrix with a small network is a challenging
problem due to the size of its search space.
A $5 \times 5$ determinant is a polynomial with $120$ monomials of degree $5$
in $25$ variables. The generic inhomogeneous polynomial of this degree
has~${25+5 \choose 5}=142506$ monomials.

From Section~\ref{sec:FeatureCrossesUniversalApproximation}, we know that it is
possible to express this multivariate polynomial perfectly with a single M-layer.
In fact, a strictly upper triangular and therefore nilpotent matrix can
achieve this. We can use this fact to accelerate
the learning of the determinant by masking out the lower triangular part
of the matrix, but we do not pursue this idea here, as we want to demonstrate
that an unconstrained M-layer is capable of learning polynomials as well.

The data consist of $n \times n$ matrices with entries sampled uniformly
between $-1$ and $1$. With this sampling, the expected value of the square of
the determinant is $\frac{n!}{3^n}$. So, we expect the square of the determinant
to be $\frac{2}{9}$ for a $3 \times 3$ matrix, and $\frac{40}{81}$ for a
$5 \times 5$ matrix. This means that an estimator constantly guessing
$0$ would have a mean square error (MSE) of $\approx0.2222$ and $\approx0.4938$
for the two matrix sizes, respectively. This provides a baseline for the
results, as a model that approximates the determinant function should
yield a smaller error.

The size of the training set consists of between $2^{10}$ and $2^{17}$
examples for the $3 \times 3$ matrices, and $2^{20}$ for the $5 \times 5$
matrices. The validation set is $25\%$ of the training set size, in
addition to it. Test sets consist of $10^6$ matrices.

The M-layer has $d = 9$ and $n$ between $6$ and $12$ for $3 \times 3$
determinants, and $n = 24$ for $5 \times 5$ determinants. The DNNs has $2$ to
$4$ equally-sized hidden layers, each consisting of $5, 10, 15, 20, 25$ or $30$
neurons, for the $3 \times 3$ matrices, and $5$ hidden layers of size $100$ for
the $5 \times 5$ matrices.

An RMSprop optimizer is used to minimize the MSE with an initial learning rate
of~$10^{-3}$, decay $10^{-6}$, and batch size $32$. These values are
chosen to be in line with those chosen in Section~\ref{sec:LearningImages}. The
learning rate is reduced by $80\%$ following $10$ epochs without validation
accuracy improvement. Training is carried for a maximum of $256$ epochs, with
early stopping after $30$ epochs without validation accuracy improvement.

\begin{figure}
\centering
\input{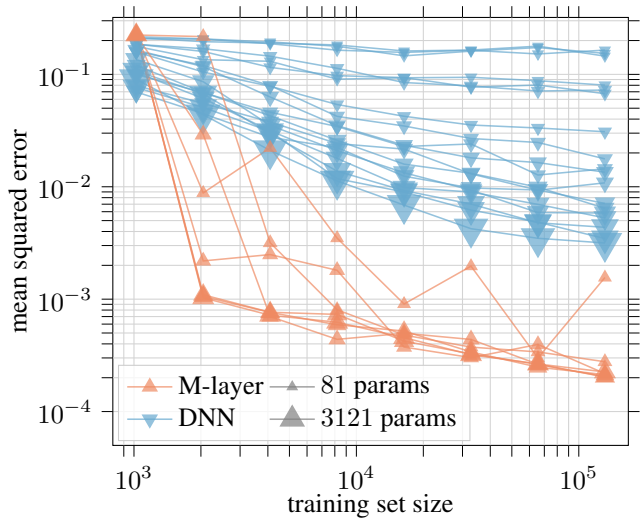}
\caption{Learning the determinant of a $3 \times 3$ matrix with M-layers and
DNNs with various parameter counts and training set sizes.}
\label{fig:error_3x3det}
\end{figure}

Figure~\ref{fig:error_3x3det} shows the results of learning the determinant of
$3 \times 3$ matrices. The M-layer architecture is able to learn from fewer
examples compared to the DNN. The best M-layer model learning on $2^{17}$
examples achieves a mean squared error of $\approx2\cdot10^{-4}$ with
$811$~parameters, while the best DNN has a mean squared error of $\approx0.003$
with $3121$~parameters.

\begin{figure}
\centering
  \begin{subfigure}[t]{0.4\textwidth}
  \newcommand\pointsdata{detscatter_dnn.dat}
 % Comment out the following line and define \pointsdata
% in main file when using standalone from main file.
% Format of data is {.2/.42, .123/.124, -.3333/-.3333},
% x-coordinate/y-coordinate.
\newcommand\loadpoints[1]{\CatchFileDef\points{#1}{\endlinechar=-1}}
\newcommand\myscale{1.3}
\newcommand\pointsize{.02*\myscale}
\loadpoints{\pointsdata}
\begin{tikzpicture}
    [scale=\myscale]

\tikzmath{\minx = -1;
          \miny = \minx;
          \maxx = 1;
          \maxy = \maxx;
          \ticklength = .1;
          \belowminy = \miny - \ticklength;
          \belowminx = \minx - \ticklength;}

\foreach \x / \y in \points{
 \node[fill,circle,inner sep=0pt,minimum size=.3pt, opacity=0.9]  at (\x,\y){};
}
\draw (\minx, \maxy) -- (\maxx, \maxy);
\draw (\maxx, \maxy) -- (\maxx, \miny);
\draw (\minx, \miny) -- node[pos=0.5,below=1.5em]{true determinant} coordinate (x axis mid) (\maxx, \miny);
\draw (\minx, \miny) -- node[pos=1.0,left=2.25em,rotate=90]{learnt determinant} coordinate (y axis mid) (\minx, \maxy);

\node[below=0.8cm] at (x axis mid) {};

\foreach \i [evaluate=\i as \x using \i] in {-1,-.5,0,.5, 1}{
  \draw (\x,\miny)-- (\x ,\belowminy)
    node[anchor=north] {\i};
  \draw (\minx,\x) -- (\belowminx,\x)
    node[anchor=east] {\i};}

\end{tikzpicture}
    \caption{DNN}
  \end{subfigure}%
  \begin{subfigure}[t]{0.4\textwidth}
   \newcommand\pointsdata{detscatter_mel.dat}
   % Comment out the following line and define \pointsdata
% in main file when using standalone from main file.
% Format of data is {.2/.42, .123/.124, -.3333/-.3333},
% x-coordinate/y-coordinate.
\newcommand\loadpoints[1]{\CatchFileDef\points{#1}{\endlinechar=-1}}
\newcommand\myscale{1.3}
\newcommand\pointsize{.02*\myscale}
\loadpoints{\pointsdata}
\begin{tikzpicture}
    [scale=\myscale]

\tikzmath{\minx = -1;
          \miny = \minx;
          \maxx = 1;
          \maxy = \maxx;
          \ticklength = .1;
          \belowminy = \miny - \ticklength;
          \belowminx = \minx - \ticklength;}

\foreach \x / \y in \points{
 \node[fill,circle,inner sep=0pt,minimum size=.3pt, opacity=0.9]  at (\x,\y){};
}
\draw (\minx, \maxy) -- (\maxx, \maxy);
\draw (\maxx, \maxy) -- (\maxx, \miny);
\draw (\minx, \miny) -- node[pos=0.5,below=1.5em]{true determinant} coordinate (x axis mid) (\maxx, \miny);
\draw (\minx, \miny) -- node[pos=1.0,left=2.25em,rotate=90]{learnt determinant} coordinate (y axis mid) (\minx, \maxy);

\node[below=0.8cm] at (x axis mid) {};

\foreach \i [evaluate=\i as \x using \i] in {-1,-.5,0,.5, 1}{
  \draw (\x,\miny)-- (\x ,\belowminy)
    node[anchor=north] {\i};
  \draw (\minx,\x) -- (\belowminx,\x)
    node[anchor=east] {\i};}

\end{tikzpicture}
    \caption{M-layer}
  \end{subfigure}
  \caption{Learning the $5 \times 5$ determinant. Each scatter plot shows
  $5000$ points, each corresponding to a pair
  \emph{(true determinant, learnt determinant)}.}
  \label{fig:5x5det}
\end{figure}
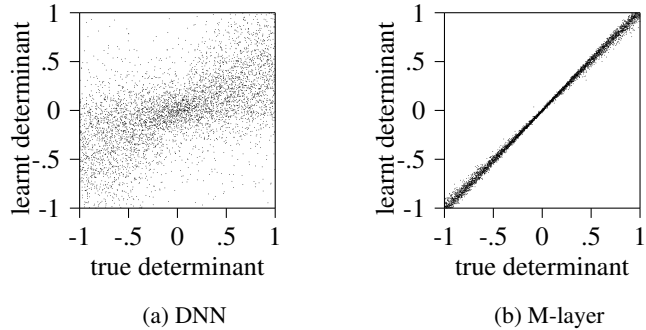

Figure~\ref{fig:5x5det} shows the results of learning the determinant of $5
\times 5$ matrices. An M-layer with $14977$ parameters outperformed a DNN
with $43101$ parameters, achieving a MSE of $0.279$ compared to $0.0012$.

\hspace{-2cm}
\begin{table*}[h]
\caption{Comparison of classification performance on image recognition
         datasets. We compare the M-layer to three general-purpose
         types of architectures: fully-connected (f.c.) networks with different
         activation functions \cite{lin2015}, NODE \cite{chen2018},
         and ANODE \cite{dupont2019}. The sources are listed as
         [D]~\cite{dupont2019} and [L]~\cite{lin2015}.}
\label{table:accuracies}
\vskip 0.15in
\begin{adjustbox}{center}
\begin{small}
\begin{sc}
\begin{tabular}{lclccr}
\toprule
  Architecture & Convolutional? & Problem & Accuracy \% (mean $\pm$ S.D.) & Parameters & Source \\
\midrule
  M-LAYER        &  no   &    MNIST      & $97.99~\pm 0.12$       & \phantom{0\,0}\num{68885} &   \\
  NODE           &  yes  &    MNIST      & $96.40~\pm 0.50$       & \phantom{0\,0}\num{84000} & [d]    \\
  ANODE          &  yes  &    MNIST      & $98.20~\pm 0.10$       & \phantom{0\,0}\num{84000} & [d]    \\
\rule{0pt}{0.15in}%
  M-LAYER        &  no   &    CIFAR-10   & $54.17~\pm 0.36$       & \phantom{0\,}\num{148965} &   \\
  SIGMOID (f.c.) &  no   &    CIFAR-10   & $46.63\phantom{~\pm 0.00}$              & \num{8049010} & [l]  \\
  ReLU (f.c.)    &  no   &    CIFAR-10   & $56.29\phantom{~\pm 0.00}$              & \num{8049010} & [l]  \\
  PReLU (f.c.)   &  no   &    CIFAR-10   & $51.94\phantom{~\pm 0.00}$              & \num{8049010} & [l]  \\
  Maxout (f.c.)  &  no   &    CIFAR-10   & $52.80\phantom{~\pm 0.00}$              & \num{8049010} & [l]  \\
  NODE           &  yes  &    CIFAR-10   & $53.70~\pm 0.20$       & \phantom{0\,}\num{172000} & [d]   \\
  ANODE          &  yes  &    CIFAR-10   & $60.60~\pm 0.40$       & \phantom{0\,}\num{172000} & [d]   \\
\rule{0pt}{0.15in}%
  M-LAYER        &  no   &    SVHN       & $81.19~\pm 0.23$       & \phantom{0\,}\num{148965} &   \\
  NODE           &  yes  &    SVHN       & $81.00~\pm 0.60$       & \phantom{0\,}\num{172000} & [d]   \\
  ANODE          &  yes  &    SVHN       & $83.50~\pm 0.50$       & \phantom{0\,}\num{172000} & [d]   \\
\bottomrule
\end{tabular}
\end{sc}
\end{small}
\end{adjustbox}
\vskip -0.1in
\end{table*}

\subsection{Learning Image Datasets}
\label{sec:LearningImages}

We assess the performance of the M-layer on three image classification tasks:
MNIST \cite{lecun1998gradient}, CIFAR10 \cite{krizhevsky2009}, and SVHN
\cite{yuval2011}.

The following procedure is used for all M-layer experiments in this section.
The training set is randomly shuffled and $10\%$ of the shuffled data is set
aside as a validation set. The M-layer dimensions are $d=35$ and $n=30$, which
are chosen by a random search in the interval $[1, 100]$.  An SGD optimizer
is used with initial learning rate of~$10^{-3}$, momentum $0.9$, and batch
size $32$. The learning rate is chosen as the largest value that gave a stable
performance, momentum is fixed, and the batch size is chosen as the
best-performing in $(32, 64)$.  The learning rate is reduced by $80\%$
following $5$ epochs without validation accuracy improvement. Training is
carried for a maximum of $150$ epochs, with early stopping after $15$ epochs
without validation accuracy improvement. The model that performs best on the
validation set is tested. Accuracy values are averaged over at least $30$
runs.

We compare the performance of the M-layer with three recently-studied
general-purpose architectures. As the M-layer is a novel architecture and
no additional engineering is performed to obtain the results in addition to
the regularization process described above, we only compare it to other generic
architectures that also use no architectural modifications to improve their
performance.

The results are shown in Table~\ref{table:accuracies}.
The M-layer outperforms multiple fully-connected architectures (with
sigmoid, parametric ReLU, and maxout activations), while employing significantly
fewer parameters. The M-layer also outperforms the NODE network, which is based
on a convolutional architecture. The networks that outperform the M-layer are
the ReLU fully-connected network, which has significantly more parameters,
and the ANODE network, which is an improved version of NODE and is also
based on a convolutional architecture.

Computing a matrix exponential may seem computationally demanding. To
investigate this, we compare the training time of an M-layer with that of a
DNN with similar number of parameters.  Table~\ref{tab:runtimes} shows that the
M-layer only takes approximately twice as much time to train.

\begin{figure}
  \centering
  \begin{tikzpicture}
\definecolor{color0}{rgb}{0.403921568627451,0.662745098039216,0.811764705882353}
\begin{axis}[
  ybar,
  width=7cm,
  height=4cm,
  grid style={white!82.7450980392157!black},
  xtick=,
  ymin=0,
  ymajorgrids,
  xmajorgrids,
  xticklabel=
    {$10^{\pgfmathprintnumber\tick}$},
  xlabel={max $L_\infty$ perturbation},
  ylabel={number of examples}
]
\addplot+[hist={data=x,bins=40}, color0, fill opacity=0.2, semithick] file {robustness.dat}; % already logscale

\end{axis}
\end{tikzpicture}
\caption{Maximum $L_\infty$ perturbation on the correctly classified CIFAR-10
  test samples that is guaranteed not to produce a misclassification.}
  \label{fig:robustness-histo}
\end{figure}
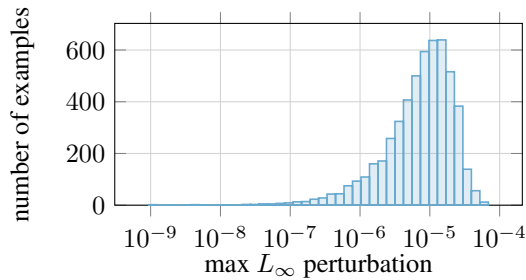

We also compute the robustness bounds of the M-layer trained on CIFAR-10, as
described in Section~\ref{sec:Robustness}. We train~$n=20$ models with
$\delta_{in} \approx 200$, $\lVert S \rVert_2 \approx 3$, and $\lVert M \rVert_2$
typically a value between $3$ and $4$. The maximum $L_2$ variation of
the vector of accumulated evidences is $\approx 1$. This results in a
typical $L_\infty$ bound for robustness of $\approx 10^{-5}$ on the whole set
of correctly classified CIFAR-10 test samples. In comparison, an analytical
approach to robustness similar to ours \cite{peck2017lower},
which uses a layer-by-layer analysis of a traditional DNN, achieves $L_2$
bounds of~$\approx 10^{-9}$.
Figure~\ref{fig:robustness-histo} shows the distribution of $L_\infty$ bounds
obtained for the M-layer.

\begin{table}[h!]
\caption{Comparison of training time per epoch for CIFAR-10 on a Nvidia V100
  GPU. The M-layer dimensions are $d = 35$ and $n = 30$. The DNN has $4$
  layers of size $43-100-100-10$.} \label{tab:runtimes}
\begin{center}
\vskip 0.15in
\begin{small}
\begin{sc}
\begin{tabular}{lcr}
\toprule
Architecture & Parameters & \multicolumn{1}{c}{Training time}\\
\midrule
M-Layer  & $148965$ & $8.67s~\pm 0.56$ \\
ReLU DNN & $147649$ & $4.12s~\pm 0.26$ \\
\bottomrule
\end{tabular}
\end{sc}
\end{small}
\end{center}
\vskip -0.1in
\end{table}

Early experiments show promising results on the same datasets when
applying advanced machine learning techniques to the M-layer, such as combining
the M-layer with convolutional layers and using dropout for regularization.
As the scope of this paper is to introduce the basics of this architecture,
we defer this study to future work.

\section{Conclusion}
\label{sec:Conclusion}

This paper introduces a novel model for supervised machine learning based on
a single matrix exponential, where the matrix to be exponentiated
depends linearly on the input. The M-layer is a powerful yet mathematically
simple architecture that has universal approximator properties and
that can be used to learn and extrapolate several problems that traditional
DNNs have difficulty with.

An essential property of the M-layer architecture is its natural ability to
learn input feature crosses, multivariate polynomials and periodic functions.
This allows it to extrapolate learning to domains outside the training data.
This can also be achieved in traditional networks by using specialized units
that perform custom operations, such as multiplication or trigonometric
functions. However, the M-layer can achieve this with no additional
engineering.

In addition to several mathematical benchmarks, we have shown that the M-layer
performs competitively on standard image recognition datasets when compared to
non-specialized architectures, sometimes employing substantially fewer
parameters.  In exchange for the benefits it provides, the M-layer only takes
around twice as much time as a DNN with the same number of parameters to train,
while also considerably simplifying hyperparameter search.

Finally, another desirable property of the M-layer is that it allows closed-form
robustness bounds, thanks to its powerful but relatively simple mathematical
structure.

We provide source code in TensorFlow that can be used to train and further
explore the capabilities of the M-layer. Future work will focus on adapting the
M-layer for specialized tasks, such as hybrid architectures for image
recognition, and advanced regularization methods inspired by the connection
between the M-layer and Lie groups.
\appendix
\section{Appendix}
\subsection{Universal Approximation Theorem}

We show that a single M-layer model that uses sufficiently large
matrix size is able to express any polynomial in the input
features. This is true even when we restrict the matrix to be
exponentiated to be nilpotent or, more specifically, strictly upper
triangular. So, for classification problems, M-layer architectures are a
superset of multivariate polynomial classifiers, where matrix size
constrains the complexity of the polynomial.

\begin{theorem}[Expressibility of polynomials]
\label{thm:exprpoly}
Given a polynomial $p(x_1, \dots, x_n)$ in $n$ variables, we can choose
weight tensors for the M-layer such that it computes $p$ exactly.
\end{theorem}
\begin{proof}
  The tensor contraction applied to the result of matrix exponentiation can
  form arbitrary linear combinations, and is therefore able
  to compute any polynomial given a matrix that contains the constituent
  monomials up to constant factors. Thus, it suffices to prove that we can
  produce arbitrary monomials in the exponentiated matrix.

  Given a monomial $m$ of degree $d-1$, we consider the $d\times d$ matrix $U$,
  that has the $d-1$ (possibly repeated) factors of the monomial on the first
  upper diagonal and zeros elsewhere. Let us consider powers of $U$. It can be
  shown that all elements of $U^i$ are equal to $0$, except for the $i$-th
  upper diagonal, and that the value in the ($d-1$)-th upper diagonal of
  $U^{d-1}$, which contains only one element, is the product of the entries of
  the first upper diagonal of $U$. This is precisely the monomial $m$ we started
  with.
  By the definition of the exponential of the matrix, $\exp(U)$ then
  contains $\frac{m}{(d-1)!}$, which is the monomial up a constant factor.

  Given a polynomial $p$ constiting of $t$ monomials, for each $1\leq i\leq t$,
  we form matrices $U_i$ for the corresponding monomial $m_i$ of $p$, as
  described above. Then we build the diagonal block matrix
  $U = \diag(U_1, \dots, U_t)$. It is clear that
  $\exp(U) = \diag(\exp(U_1), \dots, \exp(U_k))$, so we can find all monomials
  of $p$ in $\exp(U)$.
\end{proof}
To illustrate the proof, we look at a monomial $m=abcd$.
\def\arraystretch{1.2}
\begin{align*}
\begin{array}{rcl}
U &=& \left(\begin{array}{rrrrr}
0 & a & 0 & 0 & 0 \\
0 & 0 & b & 0 & 0 \\
0 & 0 & 0 & c & 0 \\
0 & 0 & 0 & 0 & d \\
0 & 0 & 0 & 0 & 0
\end{array}\right) \\
~&~&~\\
\exp(U) &=& \left(\begin{array}{rrrrr}
1 & a & \frac{1}{2}ab & \frac{1}{6}abc & \frac{1}{24}abcd \\
0 & 1 & b & \frac{1}{2}bc & \frac{1}{6}bcd \\
0 & 0 & 1 & c & \frac{1}{2}cd \\
0 & 0 & 0 & 1 & d \\
0 & 0 & 0 & 0 & 1
\end{array}\right)
\end{array}
\end{align*}

The M-layers constructed here only make use of nilpotent
matrices. When using this property as a constraint, the size
of the M-layer can be effectively halved in the implementation.

The construction from Theorem~\ref{thm:exprpoly} can be adapted to
express not only a multivariate polynomial, i.e.\ a function to $\mathbb{R}^1$,
but also functions to $\mathbb{R}^k$, which restrict to a polynomial in each
coordinate. This, together with the Stone-Weierstrass theorem
\cite{stone1948generalized}, implies the following:
\begin{corollary}
  \label{thm:polys} For any continuous function ${f\colon [a, b]^n \rightarrow
  \mathbb{R}^m}$ and any $\epsilon > 0$, there exists an M-layer model that
  computes a function $g$ such that $|f(x_0, \dots, x_{n-1})_j - g(x_{0},
  \dots, x_{n-1})_j|<\epsilon$ for all $0\leq j \le m$.
\end{corollary}

\subsubsection{Optimality of construction}

While our proof is constructive, we make no claim that the size of the matrix
used in the proof is optimal and cannot be decreased. Given a multivariate
polynomial of degree $d$ with $t$ monomials, the size of the matrix we
construct would be $t(d+1)^2$. In fact, by slightly adapting the construction,
we can obtain a size of matrix that is $td^2 +1$. Given that the total number
of monomials in polynomials of $n$ variables up to degree $d$ is
$\binom{n+d}{d}$, it seems likely possible to construct much smaller M-layers
for many polynomials. Thus, one wonders what is, for a given polynomial, the
minimum matrix size to represent it with an M-layer.

As an example, we look at the determinant of a $3\times 3$-Matrix. If the matrix
is
\[\left(\begin{array}{rrr}
a & b & c \\
d & e & f \\
g & h & i
\end{array}\right),\] then the determinant is the polynomial
$aei - afh  - bdi + bfg  + cdh - ceg$.
From Theorem \ref{thm:exprpoly}, we know that it is possible to express this
polynomial perfectly with a single M-layer. However, already an M-layer of size
$8$ is sufficient to represent the determinant of a $3 \times 3$ matrix: If
\begin{equation}
M = \left(\begin{array}{rrrrrrrr}
0 & 0 & i & f & 0 & 0 & 0 & 0 \\
0 & 0 & h & e & 0 & 0 & 0 & 0 \\
0 & 0 & 0 & 0 & 2d & -2f & 0 & 0 \\
0 & 0 & 0 & 0 & -2g & 2i & 0 & 0 \\
0 & 0 & 0 & 0 & 0 & 0 & 3c & -3b \\
0 & 0 & 0 & 0 & 0 & 0 & 3a & 0 \\
0 & 0 & 0 & 0 & 0 & 0 & 0 & 0 \\
0 & 0 & 0 & 0 & 0 & 0 & 0 & 0
\end{array}\right)
\end{equation}
then $\exp(M)$ is
\[\left(\begin{smallmatrix}
1 & 0 & i & f & -fg + di & 0 & -cfg + cdi & bfg - bdi \\
0 & 1 & h & e & -eg + dh & -fh + ei & -ceg + aei + cdh - afh & beg - bdh \\
0 & 0 & 1 & 0 & 2d & -2f & 3cd - 3af & -3bd \\
0 & 0 & 0 & 1 & -2g & 2i & -3cg + 3ai & 3bg \\
0 & 0 & 0 & 0 & 1 & 0 & 3c & -3b \\
0 & 0 & 0 & 0 & 0 & 1 & 3a & 0 \\
0 & 0 & 0 & 0 & 0 & 0 & 1 & 0 \\
0 & 0 & 0 & 0 & 0 & 0 & 0 & 1
\end{smallmatrix}\right),\]
the sum of $\exp(M)_{0,7}$ and $\exp(M)_{1,6}$ is exactly this determinant.
The permanent of a $3\times3$ matrix can be computed with an almost identical
matrix, by removing all minus signs.

\clearpage
\bibliography{intelligent_matrix_exponentiation.bib}
\bibliographystyle{alpha}

\end{document}